# BEYOND NEAREST NEIGHBOR INTERPOLATION IN DATA AUGMENTATION

*Olivier Rukundo*

Department of Electronic and Computer Engineering,
University of Limerick
e-mail: olivier.rukundo@ul.ie

## ABSTRACT

Avoiding the risk of undefined categorical labels using nearest-neighbor interpolation overlooks the risk of exacerbating pixel-level annotation errors in augmented training data. Additionally, the inherent low-pass filtering effects of interpolation algorithms exacerbate the risk of degrading high-frequency structural details within annotated regions of interest. To avoid these risks, the author modified convolutional neural networks' data transformation functions by incorporating a modified geometric transformation function, removing reliance on nearest-neighbor interpolation, and integrating a mean-based class filtering mechanism to handle undefined categorical labels with alternative interpolation algorithms. The author also implemented an offline data augmentation pipeline to generate interpolation-specific augmented training data, enabling quantitative assessment of interpolation-specific low-pass filtering effects on augmented training data. Experimental evaluation on three medical image segmentation datasets and the XBAT+ datasets demonstrated performance gains across multiple quantitative metrics.

***Index Terms***— *interpolation, data augmentation, offline data augmentation, augmented training data, convolutional neural network, geometric transformation, pixel-level annotation, semantic segmentation, object detection*

## 1. INTRODUCTION

In modern deep learning pipelines, models are commonly exposed to more augmented training data than original training data during the entire training process [1], [2], [3]. Training data augmentation requires the use of techniques that operate at the data level rather than at the model architectural level [4]. In semantic segmentation and object detection with deep learning, data augmentation increases the size of original training data to improve model generalization and predictive performance [4]. As of today, data augmentation approach still relies on interpolation algorithms to geometrically transform original training data into augmented training data. Specifically, predefined nearest-neighbor interpolation (NEA or NN) is routinely used to avoid the risk of undefined categorical labels after mask transformations. This non-extra pixel interpolation (NPI) category algorithm is routinely implemented in built-in MATLAB and Python libraries for handling masks [5], [6], [7]. On the other hand, extra pixel interpolation (EPI) algorithms, including traditional bilinear and bicubic interpolation algorithms, are traditionally used to handle images during geometric image transformations. However, there is no justification for neglecting the risk of pixel-level annotation error exacerbation, in masks, when both pixel-level annotation error exacerbation and undefined categorical label presence risks can be addressed simultaneously. Also, there is no justification for neglecting investigation into the extent of structural detail degradation within annotated regions of geometrically transformed images. Note that, the inherent flaw of nearest neighbor interpolation's reliance on rounding functions leads to heavy jagged edge artifacts [7], [8], which result in mismatches between class boundaries and object boundaries, once used for approximation purposes, ultimately, exacerbating pixel-level annotation errors [8], [10], [11]. Also, note that, completely avoiding pixel-level annotation errors is inherently challenging, as not all edge lines are defined by sharp transitions, and annotation tools still rely on manual contour refinement to achieve pixel-perfect object annotations [9]. Even when using bounding boxes or polygon annotations, there is no guarantee that the high-frequency structural and contrast information within the annotated regions will remain preserved after geometric transformation–based EPI algorithms. Furthermore, interpolation algorithms inherently introduce interpolation-specific low-pass filtering effects during geometric transformations, which degrade high-frequency structural details within annotated image regions. In [21], the authors highlighted that semantic segmentation annotations inherently contain uncertainty arising from the pixel-labeling process itself, thereby motivating uncertainty-aware handling of categorical labels. However, limited attention was given to how the NPI algorithm, traditionally used during handling of categorical labels, may influence negatively the performance of the segmentation model architecture or to how EPI algorithms may influence positively the model performance. In recent studies, the author demonstrated how the use of EPI algorithms, such as bicubic (BIC), bilinear (BIL), etc., for handling categorical labels could yield superior performance of the segmentation model architecture [5], [6]. However, the author did not evaluate effects of EPI algorithms in the context of data augmentation [5], [6]. A more recent study has shown that conventional data augmentation pipelines may amplify subtle annotation imperfections and implicit label noise when identical transformations were applied to both images and masks [20]. This further demonstrated that label-aware deformation strategies could mitigate structural distortions and preserve segmentation integrity during augmentation [20]. Nevertheless, this study did not investigate the EPI algorithms in handling categorical labels during geometric transformations for data augmentation. This did not also investigate the need for structural detail preservation within annotated or bounding-box-covered regions of images in data augmentation pipelines. Therefore, in this work, the author introduces the use of EPI algorithms in geometric transformations for handling categorical labels (or masks in training data). Also, the author created and leveraged interpolation-specific offline augmented data to investigate the degradation of structural details within annotated or bounding-boxes-covered regions. In the following sections the author presents the materials and methods used, experiments results, discussion and conclusion.

## 2. MATERIALS AND METHODS

### *A. Datasets*

**Data for semantic segmentation task**: Three datasets were downloaded from Kaggle, a large platform hosting community-shared models, datasets, and codes repositories [12]. The datasets were chosen to ensure diversity in digital imaging modalities, image representativeness, and segmentation tasks. The first dataset consists of 55 representative histology breast light-microscopy images and their corresponding masks. The second dataset includes 55 representative Magnetic Resonance Imaging (MRI) brain images and corresponding masks [13]. Lastly, the third dataset contains fifty-five representative colon polyp images, and their corresponding masks obtained through capsule endoscopy [14]. The selection of 55 image-mask pairs per dataset was based on the representativeness and informative value of the samples, rather than on dataset size alone. The objective was to avoid the inclusion of redundant or low-informative images that would contribute minimally to training outcomes, while still maintaining sufficient diversity for meaningful experimentation and balancing computational resource constraints. Figure 1 shows sample image-mask pairs from the three datasets. For semantic segmentation purposes, all images and masks were resized to 256 × 256 pixels, owing to the documented advantages of using this input size for CNN-based image segmentation tasks [5]. Furthermore, each dataset was divided into training (70%), validation (15%), and test (15%) sets.

**Data for object detection task**: Two XBAT+ datasets were downloaded from the Zenodo, a general-purpose open repository built and developed by researchers, to ensure that everyone can join in Open Science [22]. Those downloaded were High-Quality Dataset (HQD) and RGB Dataset (RGBD) belonging to the first version, XBAT+ 1.0. These datasets were chosen to ensure diversity in imaging conditions and better annotation quality and size of objects of interest. Each of HQD and RGBD contained 421 images (and corresponding .txt annotation files). All these images were acquired from static images of battery-Waste Electrical and Electronic Equipment (WEEE) [23]. Figure 2 shows the example of images from HQD and RGBD. Note the XBAT 1.0 datasets are intended for battery WEEE identification, battery presence detection and battery chemistry classification [23]. In our experiments, each of the original HQD and RGBD was split into training (70%), validation (20%), and test (10%) sets. For offline data augmentation, the original training split was augmented using a common augmentation factor of two, resulting in a final or effective augmented training set for each interpolation method.

### *B. Convolutional Neural Networks*

**For semantic segmentation**: Three convolutional neural network (CNN) architectures were used: U-Net [15], SegNet [16] and DeepLabV3+ [17]. The choice of these CNNs was based on their availability in popular software libraries, especially MATLAB. Here, U-Net and SegNet architectures were used with encoder depth set to four and convolutional layer filter size set to seven (others remained in default settings). Also, DeepLabV3+ architecture, available in R2023b MATLAB, its first convolutional layer was updated to also accept a 1-channel input for grayscale training set (the rest remained in default settings).

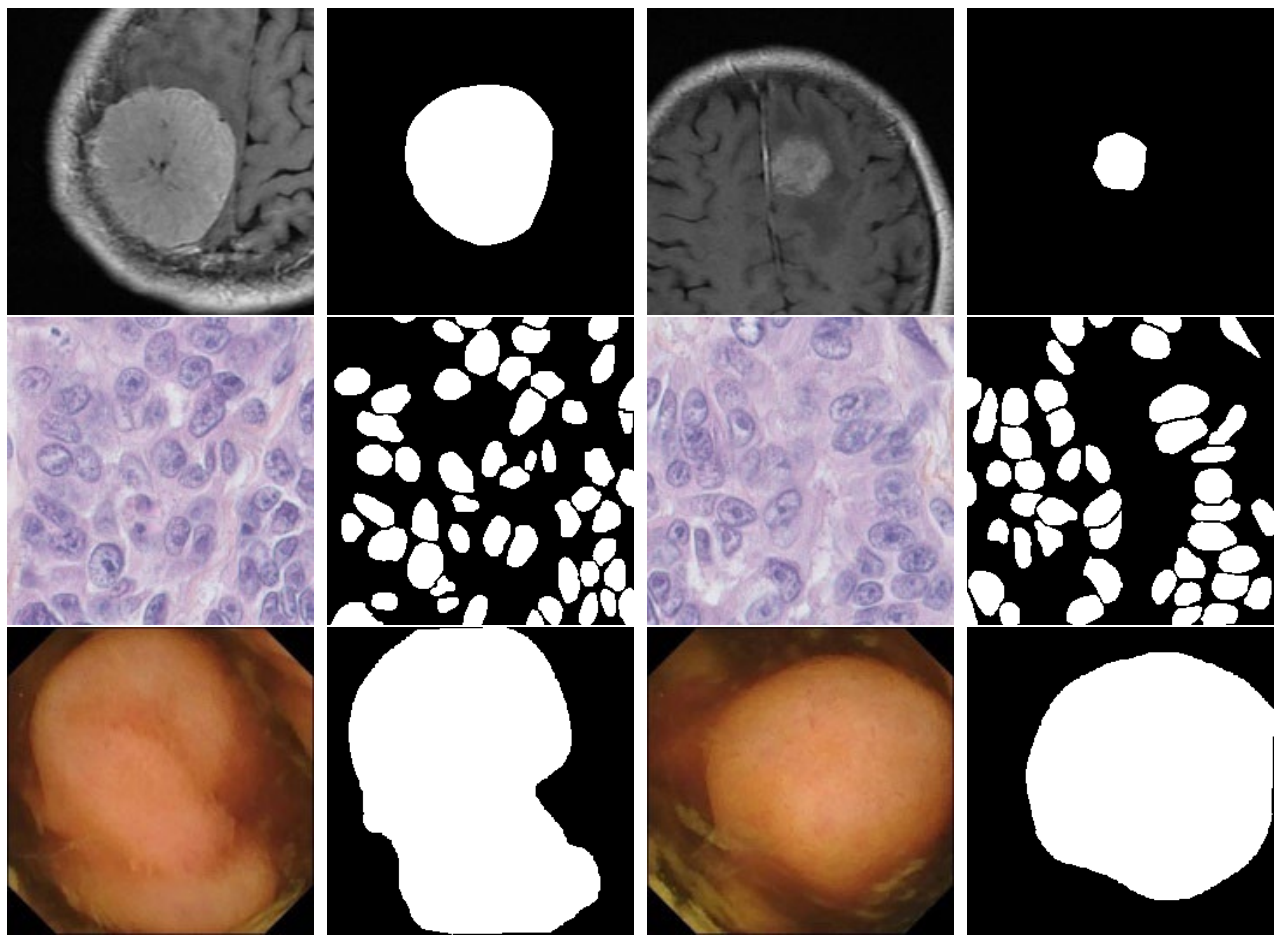

Figure 1: Top row: MRI brain image and mask pairs. Middle row: Microscopic histology breast image and mask pairs. Bottom row: Capsule endoscopy polyp image and mask pairs.

In this context, base architectures of interest, used in DeepLabV3+, were ResNet50, MobileNetV2, and ResNet18. However, experimental results obtained using ResNet50 were included in this work, primarily due to its superior performance compared to the results from other two base-architectures. Hyperparameters relevant to these CNNs' architectures were set to meet the performance requirements, specifically epochs = 30, ADAM as optimizer, initial learning rate = 0.0001, batch size = 2, L2 regularization = 0.000005. NVIDIA's GeForce RTX3070 graphic card equipped workstation was used for training purposes.

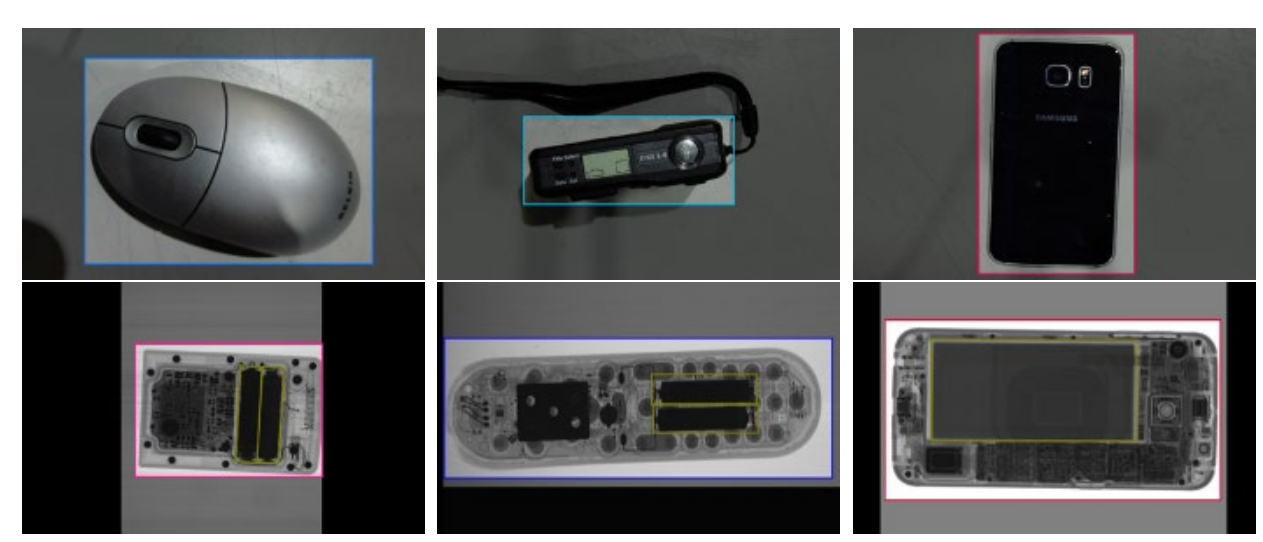

Figure 2: Top row: RGBD images with bounding box annotations around the battery-WEEE. Bottom row: HQD images with bounding box annotations around battery and battery-WEEE.

**For object detection**: The most recent YOLO version, YOLO26 (You Only Look Once 26) was used [24], [25]. The implementation and training of the YOLO26 object detection model were performed on a SLURM-managed High-Performance Computing (HPC) cluster belonging to the University of Limerick and equipped with NVIDIA A100-SXM4-40GB GPUs. GPU resource utilization and system performance were monitored using NVIDIA's nvidia-smi utility, which provides real-time statistics on GPU memory usage, processing load, temperature, power consumption, CUDA environment, and active computational processes throughout the training and inference stages.

*C. Proposed Functions for Categorical Data Handling*

The traditional version of the data augmentation pipeline is shown in Figure 3 (top). In this case, when it comes to handling categorical labels, this function leverages the geometric transformation function (*imwarp*) with its routinely imposed nearest neighbor interpolation algorithm. Given risks associated with the nearest neighbor interpolation mentioned earlier, a modified version, taking also into account the pixel-level annotation error exacerbation risk, was developed as shown in Figure 3 (bottom).

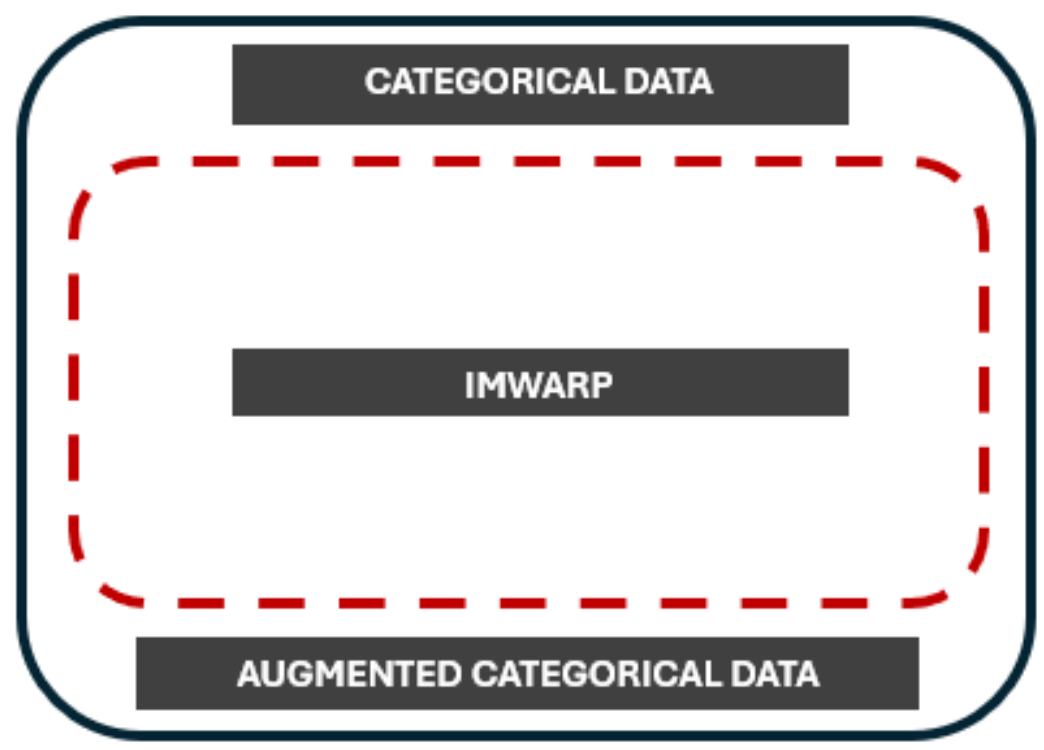

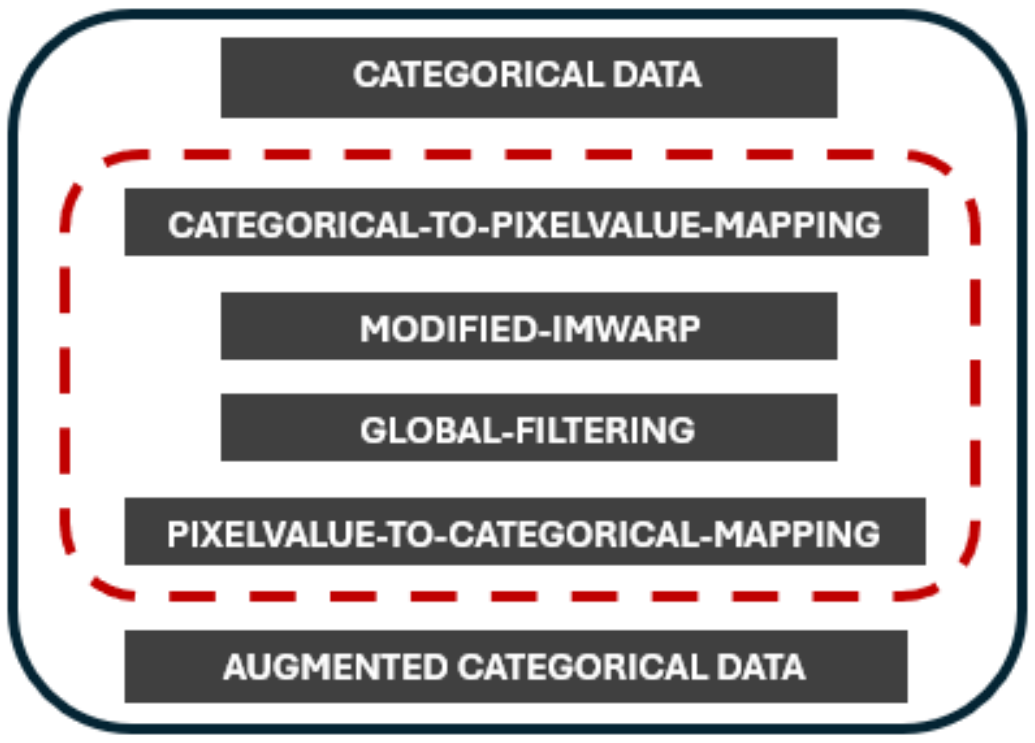

Figure 3: Comparison between the traditional and proposed functions for categorical data handling: Top: Traditional function that relies on NPI algorithms. Bottom: Proposed function that leverages EPI algorithms.

As can be seen, in Figure 3 (bottom), the mapping operations were straight forward for converting categorical labels to their corresponding pixels values or vice versa. When it comes to the modification of the imwarp function, only lines imposing the nearest neighbor interpolation algorithm were commented, to allow the use of any EPI algorithm. Due to the risk of producing undesired pixels when EPI algorithms are used for handling categorical label mapping, a filtering mechanism was proposed to remove this risk thus preventing the occurrence of undefined categorical labels at the end of the transformation (see Figure 3-bottom). In a recent study, the undesired pixel filtering mechanism was developed and applied externally before the training set content was fed to the CNN of interest, during multiclass semantic segmentation tasks [5], [6]. Here, the author internally integrated the undesired pixel filters mechanism to make the filtering process part of the augmentation process, during binary class semantic segmentation tasks. For simplicity purposes, the author developed a mean-based filtering mechanism, dubbed *global filter (GF)*, which leverages the Eq.1, Eq.2 and Eq.3 to filter the entire *modified-imwarp* function transformed mask.

$$GF(x,y) = P(x,y), \quad if\ P(x,y) = 0 \mid P(x,y) = 255 \tag{1}$$

$$GF(x,y) = 255\,, \quad if\ P(x,y) > \mu \tag{2}$$

$$GF(x,y) = 0, \quad if\ P(x,y) \leq \mu \tag{3}$$

where, 0 and 255 are two-pixel classes corresponding to each categorical label; μ is the 'mean' of all pixel values in the modified-imwarp function transformed mask (*P*); *x* and *y* are coordinates of each pixel in *P* and *GF*. It is important to note that although the current implementation was evaluated in binary segmentation settings, the filtering mechanism can be extended to multiclass scenarios by applying class-specific thresholding or label-wise filtering after interpolation. Also, note that in our previous work [5], [6], similar filtering principles were successfully applied in multiclass semantic segmentation settings. Therefore, extending the proposed GF framework to multiclass tasks would primarily require adapting the thresholding strategy to accommodate multiple categorical label values rather than relying on a single global mean threshold.

Table 1: Data augmentation transformations and parameter ranges used for semantic segmentation

| Transformation | Range |
|---|---|
| Horizontal translation | -10: 10 |
| Vertical translation | -10: 10 |
| Horizontal shear | -15: 15 |
| Vertical shear | -15: 15 |
| Rotation | -45: 45 |
| Uniform scaling | 0.8: 1.2 |
| Brightness adjustment | 0.9:0.05:1.1 |

An example of ablation analysis of the GF framework is provided in Figure 4. Here, Figure 4's 4th row images show the effects of the GF when an EPI algorithm was used to geometrically transform the image and mask of colon polyp in the modified-imwarp function. Figure 4's 3rd row images show the effects of EPI algorithm, without the GF. Figure 4's 5th row images show the effects of NPI algorithm, which does not require filtering. Now, in terms of columns, Figure 4's 1st and 2nd rows show rotated, sheared, and translated capsule endoscopy's Colon images and their corresponding ground truths, respectively. Here, the blue rectangle shows area to be cropped for better edge outline details visibility or visualization purposes. Transformations and parameter ranges used for each architecture are summarized in Table 1. These transformations were chosen because they reflect the potential for realistic structural variations in these MRI, microscopic and endoscopic images. Note that the brightness factor only applies to images, not to masks. Also, note that these ranges were chosen without specific constraints but within reasonable limits.

*D. Description of Offline Augmentation Pipeline*

An offline augmentation framework was implemented to generate interpolation-specific augmented training data. First, random affine transformations were applied using multiple interpolation methods. Annotation consistency was preserved through geometric label remapping. The augmented samples were then pre-generated and stored separately for each interpolation method, enabling controlled evaluation of interpolation-induced low-pass filtering effects across different deep learning object detection architectures. This offline data augmentation pipeline was routinely set to produce a minimum of two copies of each image in original training data. Table 2 shows the offline augmentation pipeline outcome with two augmentation epochs.

To create the final training set used in our experiments, each of original datasets consisting of 421 images and 421 annotation files was first split into training (70%), validation (20%), and test (10%) sets. Only the training set was subsequently augmented using the offline augmentation pipeline. The final training set consisted of the original training data combined with two augmented versions (i.e., final training subset = 1 original + 2 augmented samples). Consequently, augmented training data constituted the majority of the final training data, highlighting the importance of high-quality augmented data.

Table 2: Original and final dataset distributions for interpolation-specific augmentation.

| Data | Original Train/Val/Test | Augmented Train | Final Train/Val/Test |
|---|---|---|---|
| X-ray (421) | 295/84/42 | 590 | 885/84/42 |
| RGB (421) | 295/84/42 | 590 | 885/84/42 |

In this work, the augmented training data was produced for each interpolation algorithm of interest, namely: nearest (NEA), bilinear (BIL) and Bicubic (BIC). Table 3 provides the transformation names and their corresponding parameter values. Note that the number of transformations does not change the effects of number of augmentation epochs during both offline and online data augmentation. Also, note that these transformations were chosen because they reflect the potential for realistic positional orientation, and scale variations of battery-WEEE objects and batteries in these RGB (from RGBD) and grayscale X-ray images (from HQD). It should also be noted that the use of offline data augmentation does not imply that the proposed approach is incompatible with online data augmentation. Offline augmentation was adopted primarily to reduce the computational burden and training time in these experiments.

Table 3: Data augmentation transformations and parameter ranges used for object detection

| Transformation | Range |
|---|---|
| Horizontal translation | -100: 100 |
| Vertical translation | -100: 100 |
| Rotation | -45: 45 |
| Uniform scaling | 0.5: 3.0 |

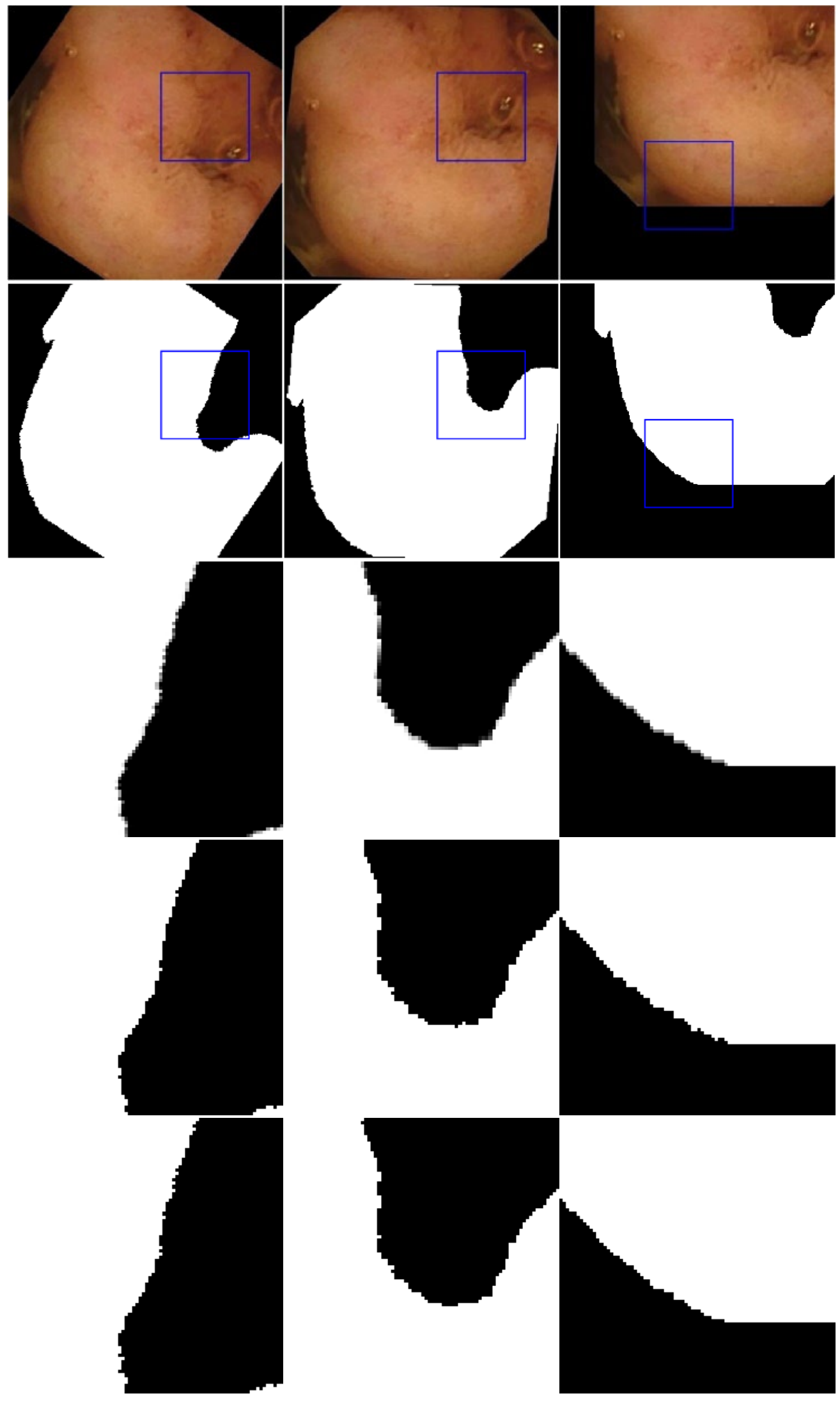

Figure 4: These three columns show rotated images and masks with cropped regions of interest. The third row shows an example of the modified-imwarp function output, while the fourth and fifth rows show the impact of GF and nearest-neighbor interpolation, respectively.

## 3. EXPERIMENTAL RESULTS

*A. Selected Evaluation Metrics*

To evaluate the performance of deep learning architectures of interest under various interpolation-specific augmented training data configurations, the following metrics were used: *Accuracy/Recall* (Eq. 4), *Intersection over Union* (*IoU)* (Eq. 5), *Precision* (Eq. 6), *BF1* (Eq.7), and *meanBFScore* (Eq. 8), *Dice score* (Eq. 9) as well as the *mAP* (Eq. 10). These metrics were chosen due to their well-established suitability for assessing segmentation and object detection with deep learning. Here, True Positive (TP) refers to predicted pixels that overlap with ground truth pixels. False Positive (FP) refers to predicted pixels that do not overlap with ground truth pixels, while False Negative (FN) refers to ground truth pixels that were not predicted. *Precision, Recall* and *Dice* scores are calculated using TP, FP, and FN [18].

$$Recall = \frac{\mathrm{TP}}{\mathrm{TP+FN}} \tag{4}$$

$$IoU = \frac{\mathrm{TP}}{\mathrm{TP+FP+FN}} \tag{5}$$

$$Precision = \frac{\mathrm{TP}}{\mathrm{TP+FP}} \tag{6}$$

$$BF1 = \frac{2\times\mathrm{Precision}*\mathrm{Recall}}{\mathrm{Recall+Precision}} \tag{7}$$

$$meanBFScore = \frac{1}{\mathrm{C}}\sum_{i=1}^{C} BF1_i \tag{8}$$

For the *meanBFScore*, $BF1_i$ is the *BF1* score for class *i*, and *C* is the number of classes. Here, each class's *Precision* and *Recall* scores are calculated separately, then averaged. Regarding the *mAP*, the mAP@50 is reported, where a predicted object is considered a true positive if its *IoU* with the ground-truth object is at least 0.5. The final mAP is computed as the mean of the Average Precision (AP) values across all classes.

$$Dice = \frac{2\times\mathrm{TP}}{2\times\mathrm{TP+FP+FN}} \tag{9}$$

$$mAP = \frac{1}{\mathrm{N}}\sum_{i=1}^{N} AP_i \tag{10}$$

*B. Quantitative Evaluation*

The idea of first using ranks of metric scores stemmed from statistical test concepts [19]. In this work, it was adopted to simplify the visualization of small differences among scores over the range of 0 to 1. Here, the highest raw score was given the rank 1, the lowest raw was given the rank 3. In other words, the small rank means the better performance. Figure 5, Figure 6, and Figure 7 present the sums of ranks computed from Accuracy, DiceScore, IoU, and meanBFScore for the BIC_BIC and NEA_NEA interpolation configurations across the Brain, Breast, and Colon datasets for UNET, SEGNET, and DEEPLABV3+, respectively. Here, NEA_NEA was used to handle both images and masks (IMAGE_MASK). While BIC_BIC was used to handle both images and masks. The other combination includes the BIL abbreviation, which means the bilinear interpolation was also used. Note that, prior experiments demonstrated that EPI_NPI configuration did not produce results superior to those of EPI_EPI. However, NPI_EPI was not previously tested but evaluated in this paper, in addition to EPI_EPI configurations. It is important to note that ranks-based scores were based on a single run experiment conducted on a Nvidia GPU equipped workstation.

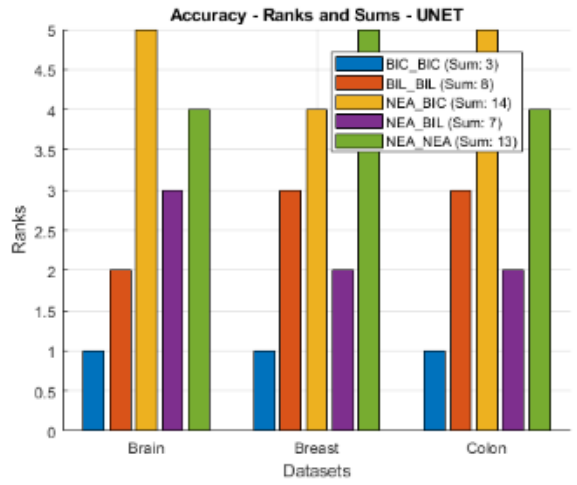

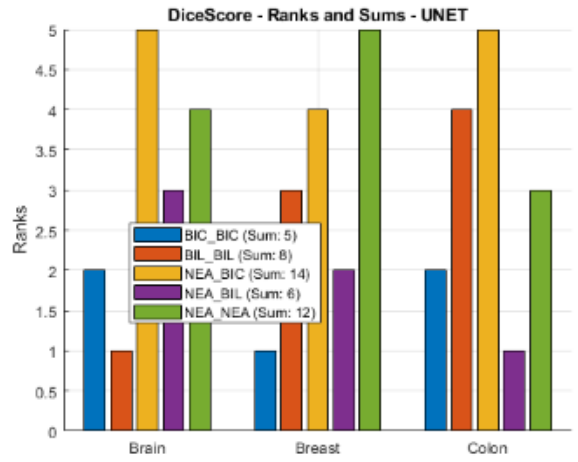

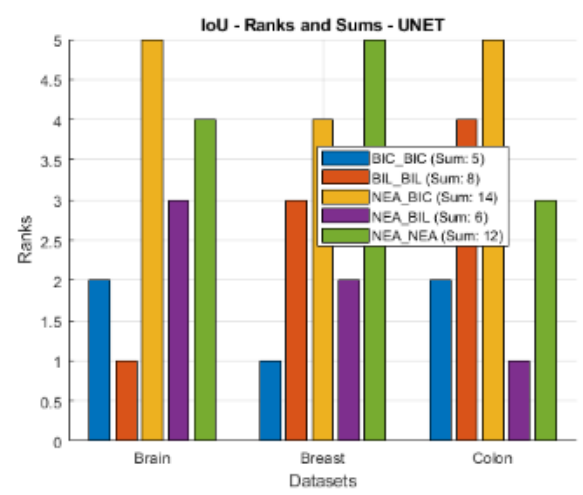

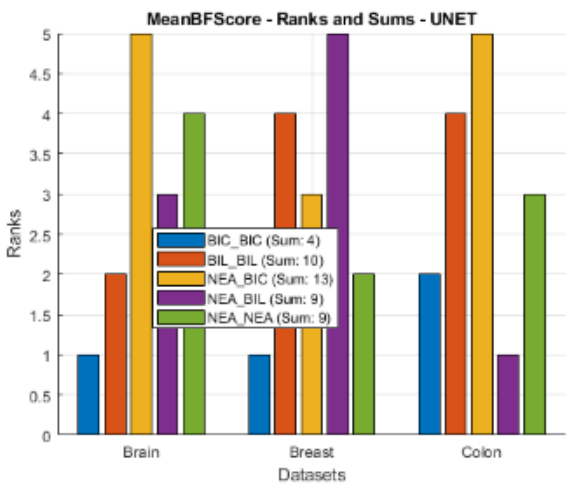

Figure 5: Unet: Accuracy, Dice, IoU and meanBFScore ranks

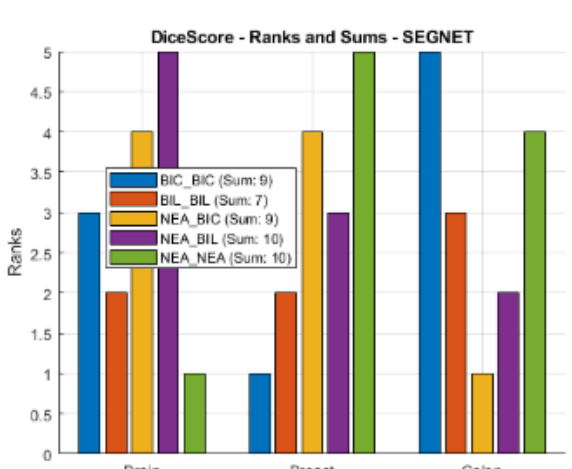

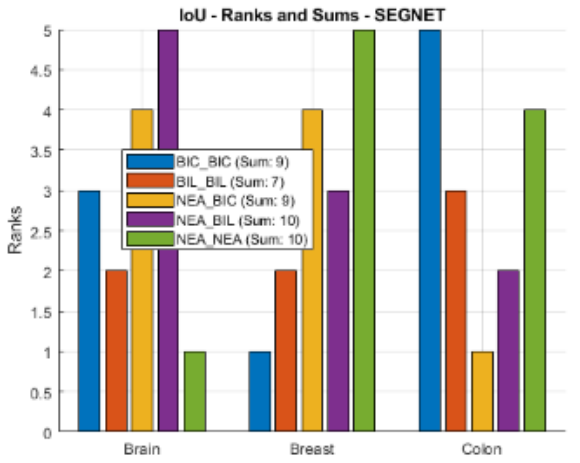

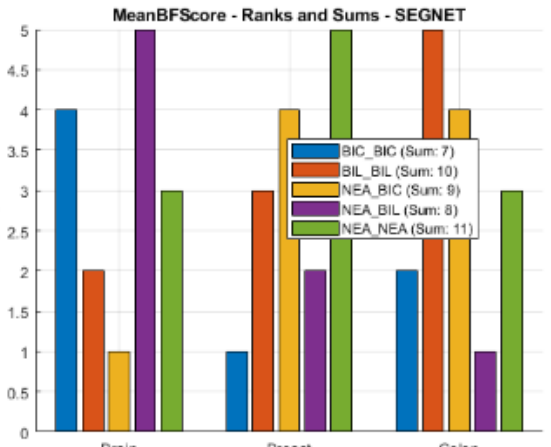

Figure 6: Segnet: Accuracy, Dice, IoU and meanBFScore ranks

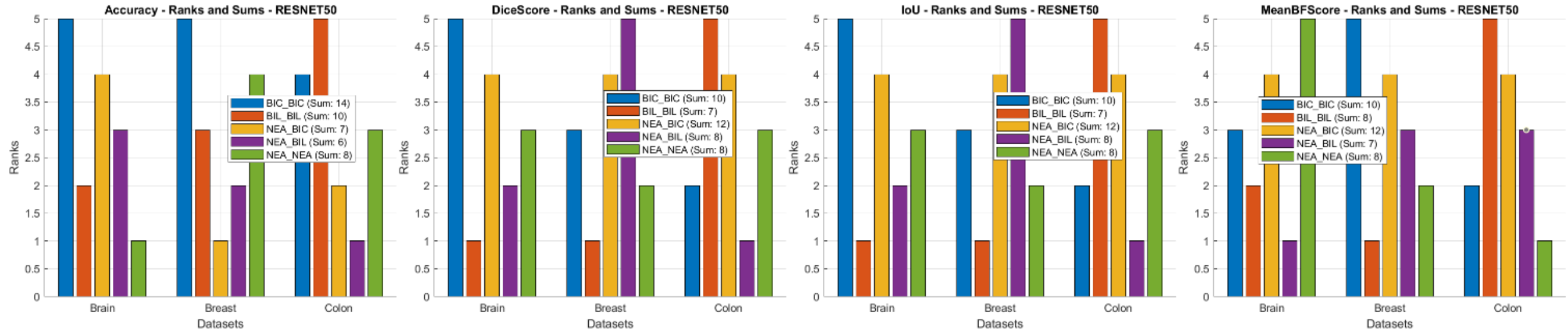

Figure 7: Deeplabv3+/R50: Accuracy, Dice, IoU and meanBFScore ranks

Table 4: Results obtained on the Brain dataset using the DeepLabV3+/ResNet-50 architecture, averaged over two runs

| Methods | Accuracy | Dice | IoU | meanBFScore |
|---|---|---|---|---|
| NEA_NEA | 0.8716±0.0184 | 0.8432±0.0614 | 0.7314±0.0918 | 0.4922±0.0754 |
| BIL_BIL | 0.8705±0.0176 | **0.8844**±0.0089 | **0.7928**±0.0143 | 0.5188±0.0211 |
| BIC_BIC | 0.8933±0.0339 | 0.8431±0.0078 | 0.7288±0.0116 | **0.5190**±0.0255 |
| NEA_BIL | 0.8527±0.0068 | 0.8582±0.0235 | 0.7520±0.0361 | 0.4548±0.0334 |
| NEA_BIC | **0.8990**±0.0114 | 0.8614±0.0044 | 0.7565±0.0067 | 0.4281±0.0011 |

Table 5: Results obtained on the Brain dataset using the Segnet architecture, averaged over two runs

| Methods | Accuracy | Dice | IoU | meanBFScore |
|---|---|---|---|---|
| NEA_NEA | **0.9046**±0.0311 | **0.7595**±0.0137 | **0.6124**±0.0178 | 0.2214±0.0198 |
| BIL_BIL | 0.7569±0.0428 | 0.7359±0.0268 | 0.5825±0.0336 | 0.2693±0.0994 |
| BIC_BIC | 0.7384±0.0339 | 0.7270±0.0155 | 0.5713±0.0191 | 0.3353±0.0383 |
| NEA_BIL | 0.7235±0.0270 | 0.6903±0.0057 | 0.5271±0.0066 | 0.3139±0.0579 |
| NEA_BIC | 0.7195±0.0285 | 0.7235±0.0111 | 0.5669±0.0136 | **0.3476**±0.0733 |

Table 6: Results obtained on the Brain dataset using the Unet architecture, averaged over two runs

| Methods | Accuracy | Dice | IoU | meanBFScore |
|---|---|---|---|---|
| NEA_NEA | 0.4894±0.0777 | 0.6231±0.0559 | 0.4538±0.0590 | 0.3179±0.0106 |
| BIL_BIL | 0.5591±0.0165 | 0.6748±0.0050 | 0.5093±0.0056 | **0.4148**±0.0022 |
| BIC_BIC | **0.5664**±0.0118 | **0.6820**±0.0115 | **0.5175**±0.0132 | 0.3742±0.0254 |
| NEA_BIL | 0.5302±0.0299 | 0.6615±0.0209 | 0.4944±0.0233 | 0.3572±0.0734 |
| NEA_BIC | 0.5360±0.0045 | 0.6650±0.0057 | 0.4982±0.0064 | 0.3505±0.0370 |

Table 7: Results obtained on the Breast dataset using the DeepLabV3+/ResNet-50 architecture, averaged over two runs

| Methods | Accuracy | Dice | IoU | meanBFScore |
|---|---|---|---|---|
| NEA_NEA | 0.7482±0.0054 | 0.7898±0.0028 | 0.6527±0.0039 | 0.7226±0.0076 |
| BIL_BIL | 0.7670±0.0036 | 0.7984±0.0046 | 0.6644±0.0064 | 0.7297±0.0015 |
| BIC_BIC | 0.7416±0.0214 | 0.7873±0.0073 | 0.6492±0.0099 | 0.7183±0.0057 |
| NEA_BIL | 0.7941±0.0041 | **0.8047**±0.0002 | **0.6733**±0.0003 | **0.7444**±0.0030 |
| NEA_BIC | **0.7973**±0.0071 | 0.7991±0.0012 | 0.6655±0.0017 | 0.7337±0.0004 |

Table 8: Results obtained on the Breast dataset using the Segnet architecture, averaged over two runs

| Methods | Accuracy | Dice | IoU | meanBFScore |
|---|---|---|---|---|
| NEA_NEA | 0.7595±0.0149 | **0.7583**±0.0103 | **0.6108**±0.0133 | **0.6415**±0.0180 |
| BIL_BIL | **0.8353**±0.0104 | 0.7020±0.0245 | 0.5411±0.0291 | 0.4949±0.0222 |
| BIC_BIC | 0.7896±0.0539 | 0.7202±0.0008 | 0.5627±0.0009 | 0.5524±0.0408 |
| NEA_BIL | 0.7763±0.0035 | 0.7561±0.0017 | 0.6078±0.0023 | 0.6165±0.0001 |
| NEA_BIC | 0.7896±0.0070 | 0.7557±0.0130 | 0.6074±0.0168 | 0.6113±0.0332 |

Table 9: Results obtained on the Breast dataset using the Unet architecture, averaged over two runs

| Methods | Accuracy | Dice | IoU | meanBFScore |
|---|---|---|---|---|
| NEA_NEA | 0.6650±0.0363 | 0.7168±0.0159 | 0.5587±0.0193 | **0.6468**±0.0019 |
| BIL_BIL | 0.7227±0.0136 | **0.7435**±0.0126 | **0.5918**±0.0159 | 0.6454±0.0020 |
| BIC_BIC | 0.6891±0.0115 | 0.7309±0.0055 | 0.5759±0.0068 | 0.6372±0.0058 |
| NEA_BIL | **0.6977**±0.0014 | 0.7352±0.0011 | 0.5812±0.0014 | 0.6429±0.0025 |
| NEA_BIC | 0.6758±0.0710 | 0.7222±0.0362 | 0.5658±0.0444 | 0.6444±0.0127 |

Table 10: Results obtained on the Colon dataset using the DeepLabV3+/ResNet-50 architecture, averaged over two runs

| Methods | Accuracy | Dice | IoU | meanBFScore |
|---|---|---|---|---|
| NEA_NEA | 0.9592±0.0022 | 0.9688±0.0012 | 0.9394±0.0022 | 0.6709±0.0134 |
| BIL_BIL | **0.9708**±0.0026 | 0.9712±0.0008 | 0.9440±0.0016 | 0.6447±0.0157 |
| BIC_BIC | 0.9650±0.0043 | **0.9722**±0.0015 | **0.9460**±0.0028 | **0.6797**±0.0227 |
| NEA_BIL | 0.9587±0.0006 | 0.9693±0.0005 | 0.9404±0.0009 | 0.6574±0.0072 |
| NEA_BIC | 0.9608±0.0028 | 0.9692±0.0014 | 0.9403±0.0027 | 0.6669±0.0073 |

Table 11: Results obtained on the Colon dataset using the Segnet architecture, averaged over two runs

| Methods | Accuracy | Dice | IoU | meanBFScore |
|---|---|---|---|---|
| NEA_NEA | 0.8871±0.0028 | 0.9323±0.0017 | 0.8732±0.0029 | 0.4689±0.0051 |
| BIL_BIL | 0.8894±0.0046 | 0.9346±0.0021 | 0.8772±0.0038 | 0.4870±0.0120 |
| BIC_BIC | **0.8988**±0.0073 | **0.9397**±0.0045 | **0.8863**±0.0080 | **0.5152**±0.0035 |
| NEA_BIL | 0.8974±0.0089 | 0.9391±0.0061 | 0.8852±0.0108 | 0.4835±0.0067 |
| NEA_BIC | 0.8884±0.0043 | 0.9334±0.0022 | 0.8752±0.0040 | 0.4645±0.0112 |

Table 12: Results obtained on the Colon dataset using the Unet architecture, averaged over two runs

| Methods | Accuracy | Dice | IoU | meanBFScore |
|---|---|---|---|---|
| NEA_NEA | 0.9316±0.0003 | 0.9553±0.0009 | 0.9144±0.0016 | 0.4633±0.0374 |
| BIL_BIL | 0.9219±0.0138 | 0.9536±0.0058 | 0.9114±0.0105 | 0.4773±0.0381 |
| BIC_BIC | 0.9289±0.0059 | 0.9557±0.0029 | 0.9152±0.0054 | 0.4765±0.0344 |
| NEA_BIL | **0.9337**±0.0063 | **0.9572**±0.0024 | **0.9179**±0.0045 | **0.5041**±0.0223 |
| NEA_BIC | 0.9281±0.0025 | 0.9532±0.0017 | 0.9105±0.0031 | 0.4443±0.0241 |

Table 4, Table 5, Table 6, Table 7, Table 8, Table 9, Table 10, Table 11, and Table 12 present the average raw scores and standard deviations computed over two runs from *Accuracy*, *DiceScore*, *IoU*, and *meanBFScore*. The results correspond to the BIC_BIC, BIL_BIL, NEA_BIL, NEA_BIC and NEA_NEA interpolation configurations across the Brain, Breast, and Colon datasets using UNET, SEGNET, and DEEPLABV3+, respectively. For object detection experiments, Table 13 and Table 14 present the average raw scores and standard deviations computed over three runs from *mAP@50*, *Precision*, and *Recall*. The results correspond to the partial BIC_, BIL_, and NEA_ interpolation configurations across multimodal imaging XBAT+1.0. datasets (i.e., HQD and RGDB) using YOLO26 object detection architecture.

*C. Qualitative Evaluation*

Figure 8, Figure 9, and Figure 10 show the segmentation masks predicted by UNET, SEGNET, and DEEPLABV3+, only for the BIC_BIC and NEA_NEA interpolation configurations across the Brain, Breast, and Colon datasets. In these figures, the first column displays the test images, while the second column contains the corresponding ground-truth masks. The third and fourth columns present the segmentation masks generated under the BIC_BIC and NEA_NEA interpolation configurations, respectively.

Figure 11 shows the YOLO26-based object detection results obtained using NEA- (*column-a*), BIL- (*column-b*), and BIC- (*column-c*) based augmented training images. The first two rows present RGBD test images along with the corresponding predictions, including detected categories/classes, the number of detected objects, and the associated detection accuracy. The bottom two rows present HQD test images together with their corresponding predictions, including detected categories, object counts, and detection accuracy.

## 4. DISCUSSIONS

**Segmentation interpretation**: For Unet, BIC_BIC is clearly superior to NEA_NEA quantitatively. It has the lowest sums of ranks across *Accuracy*, *DiceScore*, *IoU*, and *meanBFScore*. The raw-score tables also support this trend, especially for Brain and Colon. Qualitatively, however, the difference is not perfectly clean: BIC_BIC gives stronger quantitative results, but some predicted masks still show boundary roughness and small artifacts. For Segnet, BIC_BIC is generally better than NEA_NEA, but the advantage is weaker than in UNET. The rank sums favor BIC_BIC in most metrics. The qualitative results also support this: the BIC_BIC masks are more compact and closer to the ground truth, while NEA_NEA introduces clearer false-positive regions, especially in the Brain case. For Deeplabv3+/R50, the conclusion is less direct. The sum-of-ranks results favor NEA_NEA over BIC_BIC overall. However, the raw-score tables show that the best method is sometimes neither BIC_BIC nor NEA_NEA, but another configuration such as BIL_BIL, NEA_BIL, or NEA_BIC depending on dataset and metric. Qualitatively, BIC_BIC and NEA_NEA are relatively close, and neither dominates visually across all datasets. In brief, results showed that specificity of interpolation configuration has a measurable effect on segmentation performance. BIC_BIC provided the strongest overall results for UNET and SEGNET, while NEA_NEA gave the better rank-based performance for DEEPLABV3+/R50. This indicates that the impact of interpolation depends on the CNN architecture and dataset characteristics, highlighting the need for flexible interpolation choices in geometric transformations during data augmentation.

**Detection interpretation**: For RGBD, BIC case is the strongest overall interpolation method. It gives the highest *mAP@50* and *Precision*. However, BIL gives the highest *Recall*, meaning BIL case detects more objects, while BIC case produces more reliable detections. For HQD, BIC case is again the strongest overall method. It gives the highest *mAP@50* and *Precision*. However, BIL case again gives the highest *Recall*. This pattern is consistent across both RGBD and HQD. Qualitatively, Figure 11 supports the quantitative results. The BIC-based predictions appear more confident and cleaner, especially in the RGBD example, where the detected object confidence is higher than NEA and BIL cases. In the HQD example, all methods detect the main objects, but BIC case maintains strong confidence while preserving the detected regions. In brief, the interpolation method affects YOLO26 detection performance. BIC-based augmentation provides the best overall detection performance, especially for *mAP@50* and *Precision*, while BIL-based augmentation favors *Recall*. It is important to note that observed standard deviations reflect, to some extent, variations in model performance associated with the interpolation methods applied during geometric transformations. Although the underlying model architecture remained unchanged, different interpolation strategies altered the original training data characteristics, resulting in measurable variations in both segmentation and detection performance.

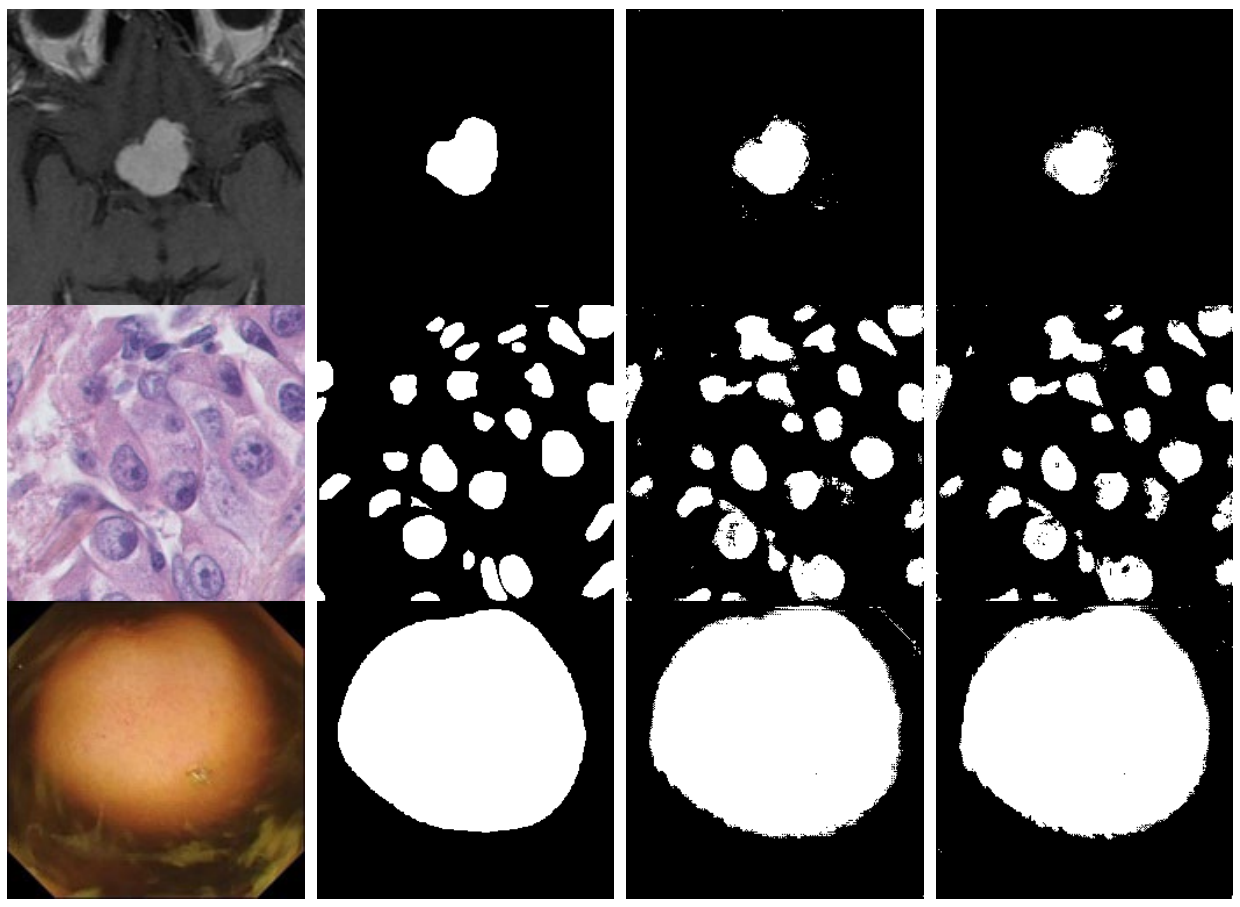

Figure 8: UNET: Left to right columns: Input image, ground truth, BC_BC predicted mask, NEA_NEA predicted mask. Top to bottom rows: Brain, Breast and Colon dataset images.

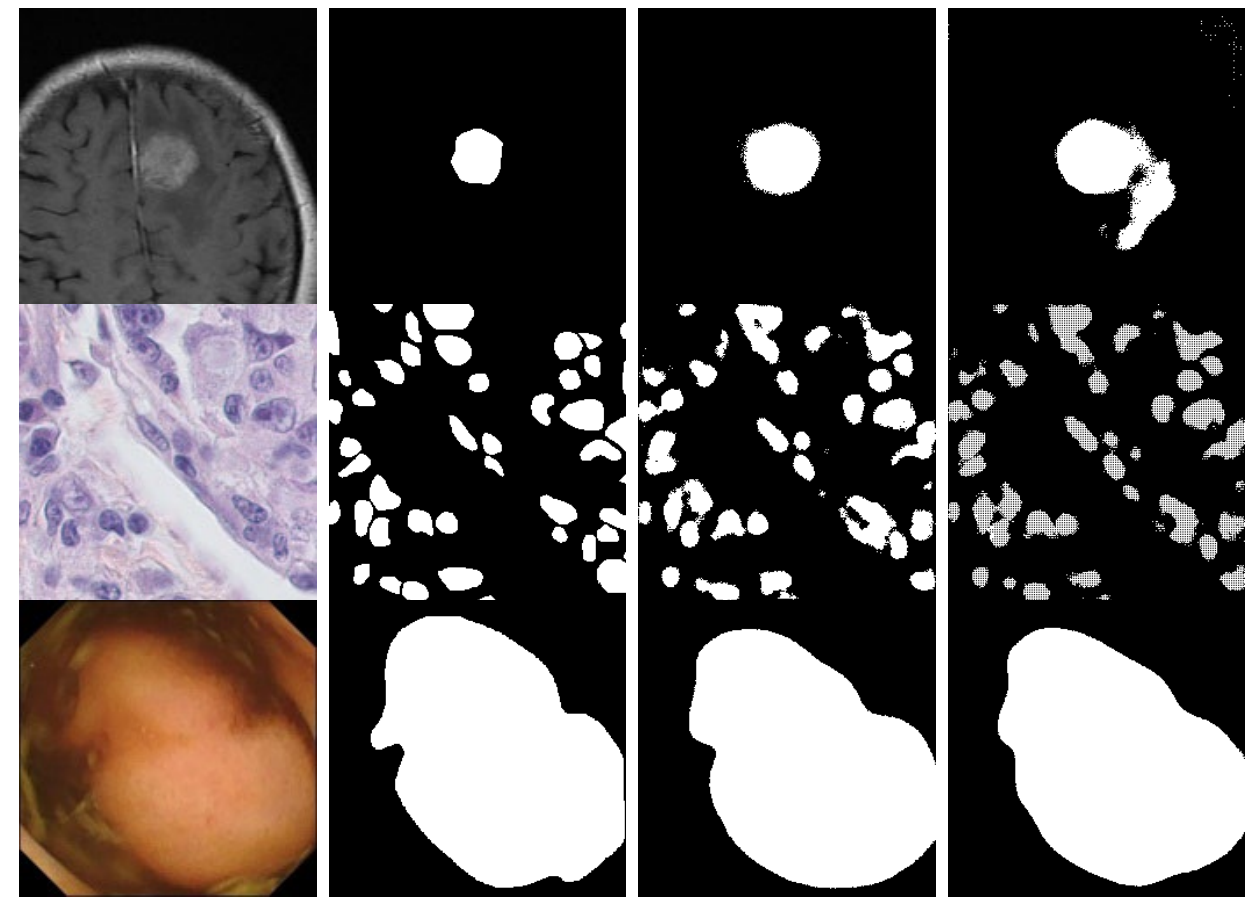

Figure 9: SEGNET: Left to right columns: Input image, ground truth, BC_BC predicted mask, NEA_NEA predicted mask. Top to bottom rows: Brain, Breast and Colon dataset images.

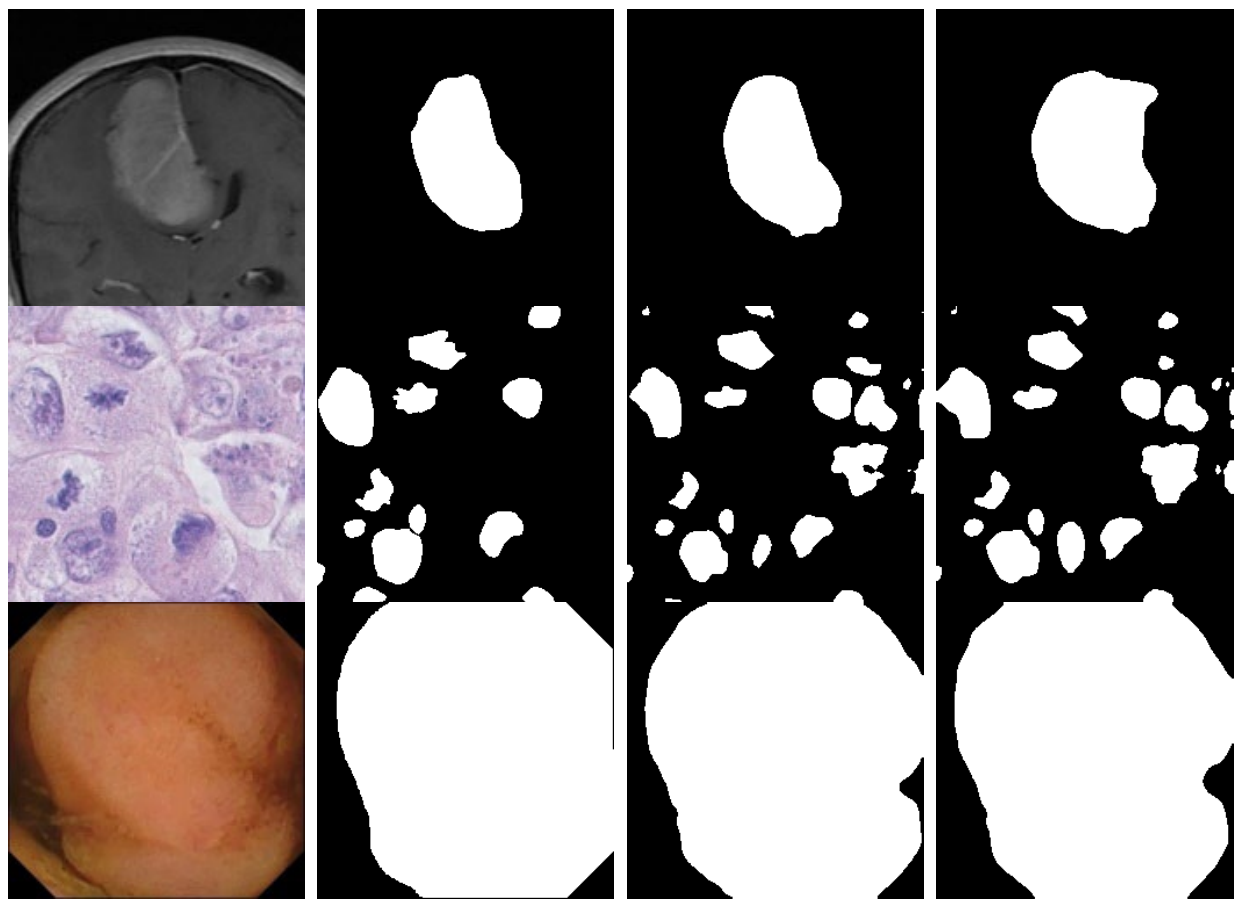

Figure 10: DEEPLABV3+/R50: Left to right columns: Input image, ground truth, BC_BC predicted mask, NEA_NEA predicted mask. Top to bottom rows: Brain, Breast and Colon dataset images.

Table 13: Average YOLO26 performance across interpolation methods on the RGBD dataset. Values are averaged over three runs.

| *Method* | *mAP@50* | *Precision* | *Recall* |
|---|---|---|---|
| NEA | 79.7 ± 5.9 | 68.1 ± 2.1 | 77.7 ± 10.3 |
| BIL | 79.7 ± 3.8 | 70.5 ± 1.0 | 82.7 ± 1.8 |
| BIC | 84.4 ± 2.2 | 80.2 ± 0.6 | 81.2 ± 6.1 |

Table 14: Average YOLO26 performance across interpolation methods on the HQD dataset. Values are averaged over three runs.

| *Method* | *mAP@50* | *Precision* | *Recall* |
|---|---|---|---|
| NEA | 78.4 ± 0.0 | 82.1 ± 0.3 | 82.4 ± 0.0 |
| BIL | 84.4 ± 2.7 | 77.0 ± 1.9 | 83.9 ± 2.7 |
| BIC | 86.8 ± 0.0 | 91.2 ± 0.0 | 79.6 ± 0.3 |

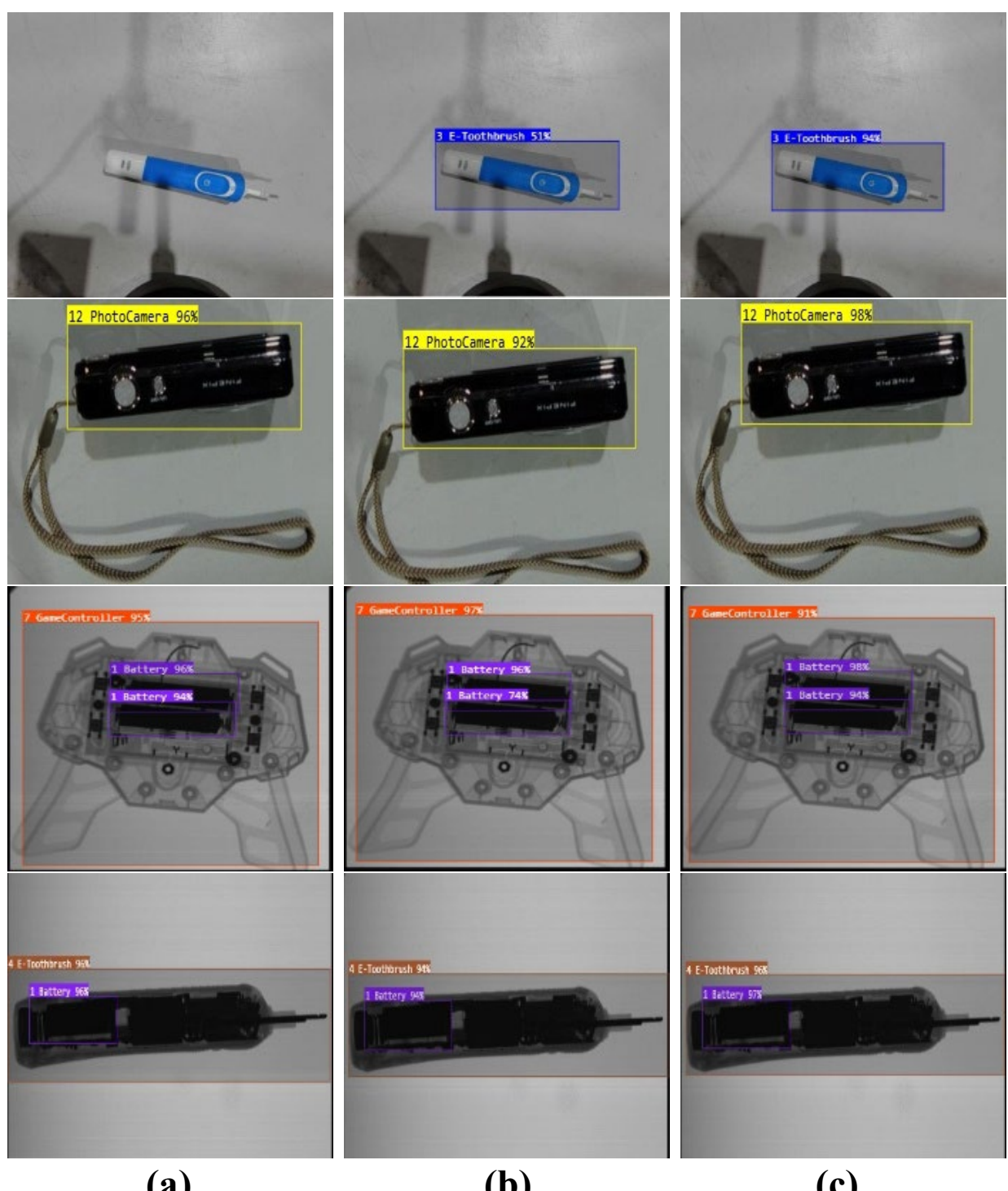

Figure 11: From left to right: (a) NEA, (b) BIL, (c) BIC-based augmented training data predictions. The top two rows present RGBD test images while the bottom two rows present the HQD test images.

## 5. CONCLUSION

This work demonstrated that flexible interpolation choices in geometric transformation-based data augmentation influence CNN-based semantic segmentation and object detection performance both quantitatively and qualitatively. Experimental results showed that EPI algorithms could better handle categorical labels, as performance mainly depends on the CNN architecture and dataset characteristics. Specifically, introducing EPI interpolation configurations improved both augmentation quality and model performance, particularly for EPI_EPI configurations. In brief, these geometric transformation-based configurations contributed to the reduction of the exacerbation of annotation errors and preservation of structural details in annotated regions. The proposed framework can also be extended to other model architectures and imaging datasets in future work.

## CONFLICT OF INTEREST

The author declares that there are no financial interests, commercial affiliations, or other potential conflicts of interest that could have influenced the objectivity of this research or the writing of this paper.

## ACKNWOLEDGEMENT

This work was supported by the University of Limerick through the XBAT+ project under the Disruptive Technologies Innovation Fund (DT 2021-0342), funded by Enterprise Ireland.

## DATA AVAILABILITY STATEMENT

Data supporting the conclusions of this paper is not made public but is available on request and approval.